\documentclass[11pt,a4paper]{article}
\usepackage{times,latexsym}
\usepackage{url}
\usepackage[T1]{fontenc}
\usepackage{bbm}
\usepackage{makecell}
\usepackage{graphicx}
\usepackage{amsmath}
\usepackage{multirow}
\usepackage{bigstrut}
\usepackage{subfigure}
\usepackage{diagbox}
\usepackage{footnote}
\usepackage{mathtools}
\usepackage{epstopdf}
\usepackage[dvipsnames,table,xcdraw]{xcolor}

\usepackage[acceptedWithA]{tacl2021v1}

\usepackage{xspace,mfirstuc,tabulary}
\usepackage{subfigure}
\usepackage{graphicx}

\newif\iftaclinstructions
\taclinstructionsfalse 
\iftaclinstructions
\renewcommand{\confidential}{}
\renewcommand{\anonsubtext}{(No author info supplied here, for consistency with
TACL-submission anonymization requirements)}
\newcommand{\instr}
\fi

\iftaclpubformat 

\else

\fi

\newcommand{\NAME}[0]{\textsc{RR}}

\usepackage{amsmath}
\DeclareMathOperator*{\argmax}{arg\,max}

\title{Rethinking with Retrieval: Faithful Large Language Model Inference}

\author{Hangfeng He$^\dag$\thanks{\, Part of this work was done while the author was at the University of Pennsylvania.}\qquad Hongming Zhang$^\ddagger$ \qquad Dan Roth$^\mathsection$ \\
  $^\dag$University of Rochester\qquad
  $^\ddagger$Tencent AI Lab, Seattle \qquad
  $^\mathsection$University of Pennsylvania\\
\texttt{hanfeng.he@rochester.edu}, \texttt{hongmzhang@global.tencent.com} \\ \texttt{danroth@seas.upenn.edu}
}

\date{}

\begin{document}
\maketitle
\begin{abstract}
Despite the success of large language models (LLMs) in various natural language processing (NLP) tasks, the stored knowledge in these models may inevitably be incomplete, out-of-date, or incorrect. This motivates the need to utilize external knowledge to assist LLMs. Unfortunately, current methods for incorporating external knowledge often require additional training or fine-tuning, which can be costly and may not be feasible for LLMs. To address this issue, we propose a novel post-processing approach, \textit{rethinking with retrieval} (\NAME{}), which retrieves relevant external knowledge based on the decomposed reasoning steps obtained from the chain-of-thought (CoT) prompting. This lightweight approach does not require additional training or fine-tuning and is not limited by the input length of LLMs. We evaluate the effectiveness of \NAME{} through extensive experiments with GPT-3 on three complex reasoning tasks: commonsense reasoning, temporal reasoning, and tabular reasoning. Our results show that \NAME{} can produce more faithful explanations and improve the performance of LLMs.\footnote{Our code is publicly available at \url{https://github.com/HornHehhf/RR}.}

\end{abstract}

\begin{figure}
	\centering
	\includegraphics[scale=0.31]{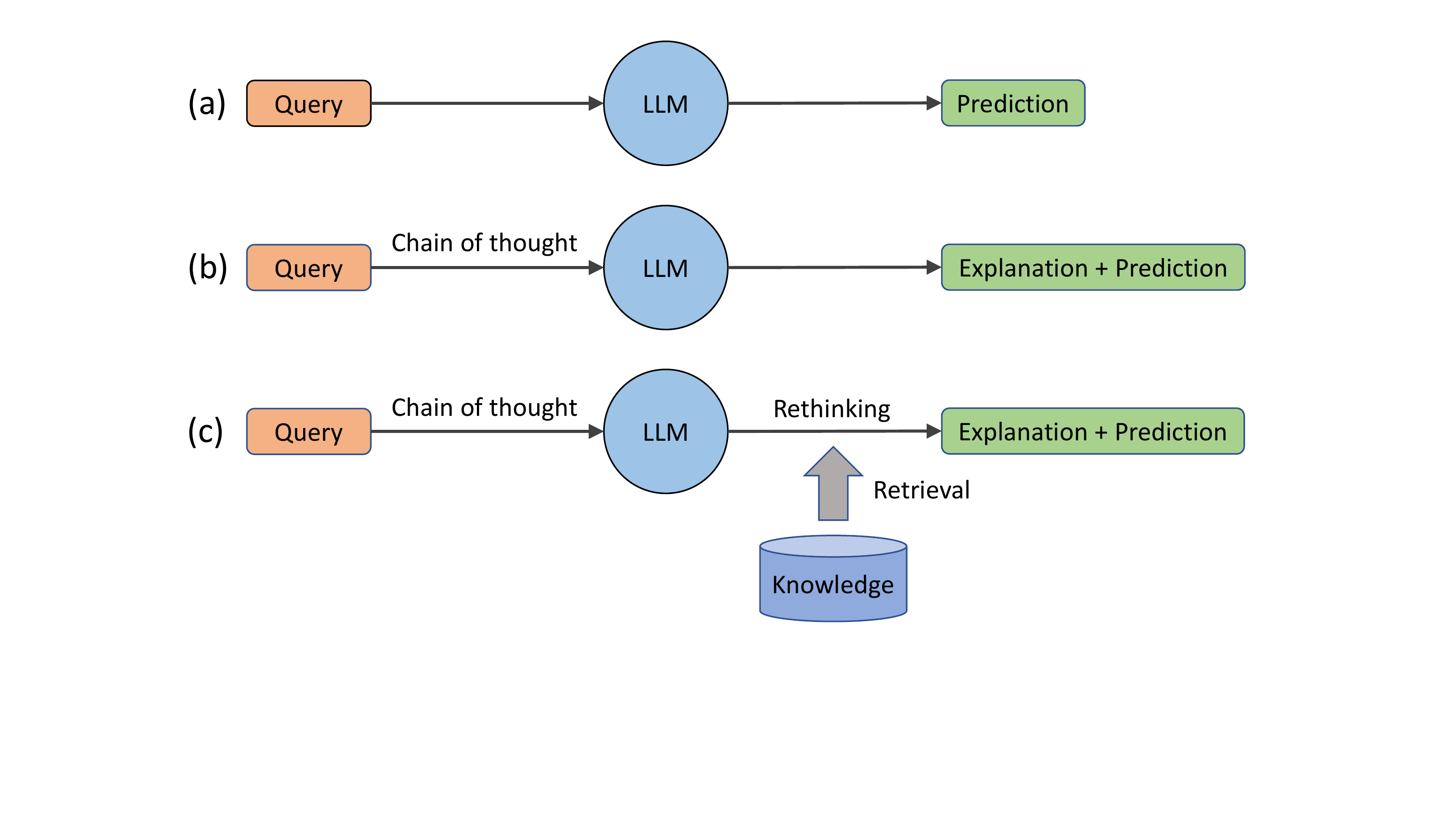}
    \caption{An overview of three approaches for using LLMs: (a) Standard prompting for generating a prediction in response to a query. (b) Chain-of-thought prompting for generating both an explanation and a prediction in response to a query. (c)  Rethinking with retrieval, our proposed approach for using the decomposed reasoning steps obtained from chain-of-thought prompting to retrieve relevant external knowledge for LLMs, leading to more faithful explanations and improved predictions in response to a query.}
	\label{fig:framework}
\end{figure}
\section{Introduction}

Large language models (LLMs) have shown exceptional performance across various tasks through in-context learning without task-specific training or fine-tuning \cite{brown2020language, chowdhery2022palm, zhang2022opt, ouyang2022training}. Recent progress in prompting \cite{wei2022chain, zhou2022least, kojima2022large} and decoding \cite{wang2022self} has made it feasible for LLMs to tackle tasks that demand complex reasoning. 

However, the knowledge stored in LLMs might inevitably be incomplete, out-of-date, or incorrect. As a result, external sources of knowledge, such as Wikipedia, may be essential for the successful deployment of LLMs for real-world applications. Previously, people tried to utilize knowledge for smaller language models (LMs), such as T5 \cite{raffel2020exploring}, BERT \cite{devlin2019bert}, and RoBERTa \cite{liu2019roberta}. However, these methods often require additional training or fine-tuning, which can be costly and thus impractical for LLMs. 

In this paper, we present a post-processing approach called \textit{rethinking with retrieval} (\NAME{}) for utilizing external knowledge in LLMs. Our method begins by using the chain-of-thought (CoT) prompting method \cite{wei2022chain} to generate a diverse set of reasoning paths, as described in \citet{wang2022self}. We then use each reasoning step in those paths to retrieve relevant external knowledge, which enables \NAME{} to provide more faithful explanations and more accurate predictions, as illustrated in Figure \ref{fig:framework}.

We evaluate the effectiveness of our proposed method, \NAME{}, on three complex reasoning tasks: commonsense reasoning, temporal reasoning, and tabular reasoning, using GPT-3 175B \cite{brown2020language} and different external knowledge sources: Wikipedia, Wikidata \cite{vrandevcic2014wikidata}, WordNet \cite{miller1995wordnet}, and Conceptnet \cite{speer2017conceptnet}. The results demonstrate that \NAME{} consistently outperforms all baselines on all three tasks without requiring additional training or fine-tuning, indicating the superiority of our approach in leveraging external knowledge to enhance the performance of LLMs. 
\section{Related Work}

\paragraph{Enhancing LMs through retrieval.} Retrieval-enhanced LMs have received significant attention as a means of improving performance through the incorporation of external knowledge. For example, the k-most similar training contexts can be retrieved to improve the estimation of the next word distribution in both the training stage \cite{borgeaud2021improving} and the inference stage \cite{khandelwal2019generalization}. Furthermore, search query generators have been adopted to generate search queries for search engines to retrieve relevant documents \cite{komeili2022internet, shuster2022language, thoppilan2022lamda}. Other approaches have utilized retrieved documents as the additional context in generation tasks \cite{joshi2020contextualized, guu2020retrieval, lewis2020retrieval}. \citet{nakano2021webgpt} instead use human feedback in a text-based web-browsing environment. Among these previous works, \citet{khandelwal2019generalization} is most closely related to our approach. However, they focus on improving local inference by using the nearest neighbor datastore constructed from training data, whereas we focus on conducting faithful inference using external knowledge. In contrast to other aforementioned approaches, which require training or fine-tuning to incorporate retrieved knowledge, we propose a post-processing method for leveraging retrieved knowledge without additional training or fine-tuning.

\paragraph{Incorporating external knowledge into LMs.} Significant effort has been devoted to leveraging external knowledge to improve the reasoning ability of LMs. Previous work has incorporated external knowledge sources such as WordNet \cite{miller1995wordnet} and ConceptNet \cite{speer2017conceptnet} to enhance LMs for tabular reasoning tasks \cite{neeraja2021incorporating, varun2022trans}. Explicit rules have also been added to inputs to improve reasoning ability over implicit knowledge \cite{talmor2020leap}. In addition, explicit knowledge from Wikidata \cite{vrandevcic2014wikidata} and implicit knowledge in LLMs have been integrated into a transformer \cite{vaswani2017attention} for visual question answering \cite{gui2021kat}. \citet{nye2021improving} instead introduces a symbolic reasoning module to improve coherence and consistency in LLMs. Among these previous works, \citet{nye2021improving} is the most relevant to our approach. Still, they focus on incorporating logical constraints to improve coherence and consistency, whereas we aim to improve the faithfulness of explanations through the use of external knowledge. In contrast to other aforementioned approaches that incorporate external knowledge before generation and require additional training or fine-tuning, our proposal leverages external knowledge in a post-processing manner to enhance LMs without additional training or fine-tuning.

\paragraph{Uncovering latent Knowledge in LLMs.} There has been a line of work exploring the knowledge hidden within LLMs for reasoning. This has included the use of careful prompting to encourage LLMs to generate explanations in the reasoning process, such as through chain of thought prompting in few-shot \cite{wei2022chain} or zero-shot \cite{kojima2022large} learning, or through the use of scratchpads for intermediate computation \cite{nye2022show}. In addition, various methods based on sampling a diverse set of reasoning paths in LLMs have been proposed, including training verifiers to judge the correctness of model completions \cite{cobbe2021training}, calibrating model predictions based on the reliability of the explanations \cite{ye2022unreliability}, and promoting self-consistency over diverse reasoning paths \cite{wang2022self}. \citet{zelikman2022star} instead iteratively bootstrap the ability of LLMs to generate high-quality rationales from a few initial examples. \citet{liu2022generated} further propose generating knowledge from LLMs, which is then used as additional input to improve commonsense reasoning. In contrast to this line of work, our proposal focuses on leveraging external knowledge to enhance LLMs, while they aim to explore the knowledge hidden within LLMs.

\section{Rethinking with Retrieval}
\label{sec:framework}

LLMs have been shown to generate incorrect supporting facts from time to time, even when they accurately capture the perspective needed to answer a question. This phenomenon highlights intrinsic issues in the way LLMs store and retrieve knowledge, including (1) the presence of out-of-date, incorrect, or missing relevant knowledge in the pre-training corpus; (2) incorrect memorization of relevant knowledge during pre-training; and (3) incorrect retrieval of relevant knowledge during the inference stage. To address these issues, we propose the use of \NAME{}, which leverages external knowledge through the retrieval of relevant information based on decomposed reasoning steps.

\paragraph{Overview.} Given a query $Q$, we utilize chain-of-thought prompting to generate a diverse set of reasoning paths $R_1, R_2, \cdots R_N$, where each reasoning path $R_i$ consists of an explanation $E_i$ followed by a prediction $P_i$. After that, we retrieve relevant knowledge $K_1, \cdots K_M$ from a suitable knowledge base $\mathcal{KB}$ to support the explanation in each reasoning path, and select the prediction $\hat{P}$ that is most faithful to this knowledge. To better illustrate our proposal, we use ``\textit{Did Aristotle use a laptop?}'' as a running example in this work.

\paragraph{Chain-of-thought prompting.}  In contrast to standard prompting, CoT prompting \cite{wei2022chain} includes demonstrations of step-by-step reasoning examples in the prompt
to produce a series of short sentences that capture the reasoning process. For instance, given the question ``\textit{Did Aristotle use a laptop?}'', CoT prompting aims to generate the complete reasoning path ``Aristotle died in 322 BC. The first laptop was invented in 1980. Thus, Aristotle did not use a laptop. So the answer is no.'' rather than simply outputs ``No.'' Empirical results show that CoT prompting significantly improves the performance of LLMs on many multi-step reasoning tasks. Therefore, we adopt CoT prompting to obtain both explanation $E$ and prediction $P$ for the query $Q$. 

\paragraph{Sampling diverse reasoning paths.} Similar to \citet{wang2022self}, we sample a diverse set of reasoning paths $R_1, R_2, \cdots R_N$ rather than only considering the greedy path as in \citet{wei2022chain}. For the question ``\textit{Did Aristotle use a laptop?}'', the potential reasoning paths can be as follows:
\begin{itemize}
    \item[($R_1$)] Aristotle died in 2000. The first laptop was invented in 1980. Thus, Aristotle used a laptop. So the answer is yes.
    \item[($R_2$)] Aristotle died in 322BC. The first laptop was invented in 2000. Thus, Aristotle did not use a laptop. So the answer is no.
    \item[($R_3$)] Aristotle died in 322BC. The first laptop was invented in 1980. Thus, Aristotle did not use a laptop. So the answer is no.
\end{itemize}

\paragraph{Knowledge retrieval.} Different knowledge bases can be used to address different tasks. For example, to address the question ``\textit{Did Aristotle use a laptop?}'', we can use Wikipedia as the external knowledge base $\mathcal{KB}$. Information retrieval techniques can be applied to retrieve the relevant knowledge $K_1, \cdots K_M$ from Wikipedia based on the decomposed reasoning steps. Ideally, we would obtain the following two paragraphs from Wikipedia for this question:
\begin{itemize}
    \item[($K_1$)] Aristotle (384–322 BC) was a Greek philosopher and polymath during the Classical period in Ancient Greece. ...
    \item[($K_2$)] The Epson HX-20, the first laptop computer, was invented in 1980. ...
\end{itemize}

\paragraph{Faithful inference.} 
The faithfulness of each reasoning path $R_i$ can be estimated using a function $f_{\mathcal{KB}}(R_i)$, which is based on relevant knowledge $K_1, \cdots, K_M$ retrieved from the knowledge base $\mathcal{KB}$. The final prediction is obtained through the application of the following inference procedure\footnote{Note that this is the basic version of faithful inference, and further variations can be found in Section \ref{subsec:variations}.}:
\begin{equation}
\hat{P} = \argmax_{P_i \in\{P_1, \cdots, P_N\}} \sum_{i=1}^N \mathbbm{1}(P_i=P) f_{\mathcal{KB}}(R_i),
\label{eq:inference}
\end{equation}
where $P_i$ denotes the corresponding prediction in the reasoning path $R_i$. This inference procedure is designed to identify the most faithful prediction $\hat{P}$ to the knowledge base among all predictions in the $N$ reasoning paths. For instance, in the running example, given reasoning paths $R_1, R_2, R_3$ and the retrieved knowledge $K_1, K_2$, the above inference procedure would output the prediction ``So the answer is no.'', as it is supported by both $R_2$ and $R_3$ and has a higher faithfulness score compared to the prediction ``So the answer is yes.'', which is only supported by $R_1$.

\section{Experiments}
\label{sec:experiments}
In this section, we present the evaluation of our proposed method, \NAME{}, on three complex reasoning tasks: commonsense reasoning, temporal reasoning, and tabular reasoning. 

\subsection{Baselines}
\label{subsec:baselines}
We compare with the following baselines.

\paragraph{Zero-shot/few-shot prompting.} In our experiments, we consider GPT-3 with standard zero-shot/few-shot prompting as baselines, following the approach described in \citet{brown2020language}, in which zero or few in-context exemplars of input-output pairs are provided in the prompt.

\paragraph{Chain-of-thought prompting.} In addition to the standard zero-shot/few-shot prompting, we also consider GPT-3 with the CoT prompting proposed in \cite{wei2022chain} as a baseline in our experiments. This approach involves feeding LLMs step-by-step reasoning examples instead of standard input-output examples.

\paragraph{Self-consistency.} In addition, we also consider self-consistency \cite{wang2022self} as a baseline in our experiments. This approach, proposed as an alternative to the naive greedy decoding used in CoT prompting \cite{wei2022chain}, involves sampling a diverse set of reasoning paths and selecting the most consistent answer by marginalizing the sampled paths. 

\subsection{Commonsense Reasoning}
\label{subsec:commonsense}

\paragraph{Dataset description.} For commonsense reasoning, we consider the StrategyQA dataset \cite{geva2021did}, which includes questions that require implicit reasoning strategies. For example, the question ``\textit{Did Aristotle use a laptop?}'' requires \textit{implicit} decomposition into reasoning steps, while the question ``\textit{Was Aristotle alive when the laptop was invented?}'' explicitly specifies the reasoning process. The StrategyQA dataset includes $2,290$ training examples, each consisting of a question (Q), a yes/no answer (A), a decomposition (D), evidence paragraphs (E), and supporting facts (F). On average, each question requires about $2.93$ reasoning steps and $2.33$ evidence paragraphs. In addition, a development set is constructed by randomly sampling $10\%$ of the training examples (i.e., $229$ examples). The answer distribution is roughly balanced, with approximately $47\%$ "yes" questions in both the training and development sets. Unless otherwise specified, the models are evaluated on the development set\footnote{As the annotations for the test set are not publicly available, we use the development set for evaluation. This allows us to perform a more comprehensive analysis.} for StrategyQA.

\begin{table*}[t]
\centering
\scalebox{1.0}{
\begin{tabular}{c|c|c|c|c}
\Xhline{2\arrayrulewidth}
 & Methods & Commonsense & Temporal & Tabular  \bigstrut[t] \bigstrut[b]   \\ \hline
 \multirow{5}{*}{GPT-3} & Zero-shot prompting & 58.08 &  28.40 & 82.00  \bigstrut[t]  \\
 & Few-shot prompting & 63.32 & 29.59  & 83.08 \\
 & Chain-of-thought prompting & 65.94  & 33.14 & 83.33 \\ 
 & Self-consistency & 73.36 & 37.28 & 84.00   \\
  & Rethinking with retrieval & {\bf 77.73} & {\bf 39.05} & {\bf 84.83} \bigstrut[b]  \\
\Xhline{2\arrayrulewidth}
\end{tabular}}
\caption{Performance of different methods using GPT-3 on three reasoning tasks.
}
\label{table:gpt3-results}
\end{table*}

\paragraph{Implementation details.} In this part, we utilize Wikipedia as the external knowledge base $\mathcal{KB}$. For each sentence in the explanation of every reasoning path, we first apply BM25 \cite{robertson2009probabilistic} to retrieve the top 10 most relevant paragraphs from Wikipedia. In particular, we use the re-implementation of the sparse retrieval BM25\footnote{We also experimented with DPR and BM25+DPR, and found that BM25 outperformed these methods in our experiments. More details can be found in Appendix \ref{subsec:retrieval-comparison}.} in \citet{karpukhin2020dense} from Pyserini \cite{Lin_etal_SIGIR2021_Pyserini}. Subsequently, we use the pre-trained MPNet model \cite{song2020mpnet} to select the most similar paragraph based on the cosine similarity between the sentence embeddings of the retrieved paragraph and the sentence. We then employ a pre-trained natural language inference (NLI) model \cite{nie2020adversarial} to obtain the entailment and contradiction scores for the sentence, treating the most similar paragraph as the premise. The faithfulness of each reasoning path is then calculated using $f_{\mathcal{KB}}(\cdot)$ based on the entailment scores, contradiction scores, and MPNet similarities of all sentences in the explanation of the reasoning path. The final prediction for each question is obtained through faithful inference (Equation \ref{eq:inference}). More details about $f_{\mathcal{KB}}(\cdot)$ can be found in Appendix \ref{subsec:faithfulness-functions}.

\subsection{Temporal Reasoning}
\label{subsec:temporal}

\paragraph{Dataset description.} In this experiment, we use the TempQuestions dataset \cite{jia2018tempquestions} to investigate temporal reasoning. This dataset includes $1,271$ temporal questions that are divided into four classes: explicit temporal, implicit temporal, temporal answer, and ordinal constraints. The questions are paired with their answers from Freebase \cite{bollacker2008freebase}. To examine the most challenging aspect of temporal reasoning, we focus on the set of \textit{implicit} temporal questions, which contain implicit temporal expressions, including free-text temporal expressions. For example, the question ``\textit{who was governor of oregon when shanghai noon was released?}'' is an implicit temporal question. To facilitate our analysis, we only consider questions with a single answer, resulting in a total of $175$ examples. Of these examples, the first $6$ are used for prompting, and the remaining $169$ are used for evaluation.

\paragraph{Implementation details.} In this part, we utilize Wikidata \cite{vrandevcic2014wikidata} as the external knowledge base $\mathcal{KB}$, as it is the largest publicly available knowledge graph, and the data from Freebase has been migrated to Wikidata. To incorporate this knowledge into our system, we apply an entity linking system\footnote{We use the spacy entity linker: \url{https://pypi.org/project/spacy-entity-linker/}. } to each sentence in the explanation of each reasoning path to identify the corresponding Wikidata pages for all entities in the sentence. Next, we extract all temporal relations from these relevant Wikidata pages and use templates to convert these temporal relations into sentences. This step generates a set of relevant knowledge sentences for each sentence in the explanation of each reasoning path. The final prediction is then obtained by applying the procedure described in Section \ref{subsec:commonsense}, in which the retrieved paragraphs are replaced with the relevant knowledge sentences from the current part.

\subsection{Tabular Reasoning}
\label{subsec:tabular}

\paragraph{Dataset description.} We consider the \textsc{INFOTABS} dataset \cite{gupta2020infotabs} for tabular reasoning, which consists of $23,738$ human-written textual hypotheses based on premises in the form of tables extracted from $2,540$ unique Wikipedia info-boxes. We focus on the development set, which includes $1,800$ hypotheses based on $200$ tables, and only consider entailed and contradictory hypotheses as it is tricky to write CoT demonstrations for neutral hypotheses. This results in a total of $1,200$ hypotheses based on $200$ tables for evaluation, with an equal number of entailed and contradictory hypotheses.

\paragraph{Implementation details.} In this part, we utilize WordNet \cite{miller1995wordnet} and ConceptNet \cite{speer2017conceptnet} as external knowledge bases. To convert tables into textual premises, we follow the same technique as in \citet{varun2022trans}. For each premise-hypothesis pair, we follow the procedure outlined in \citet{varun2022trans} to retrieve relevant word relation triples that connect the premise and hypothesis words, such as ``married''$ \xleftrightarrow{\text{RelatedTo}}$ ``spouse''. These triples are then converted into sentences using some simple templates. The resulting sentences, along with the textual premises from the tables, serve as relevant knowledge for each sentence in the explanation of each reasoning path. To obtain the final prediction, the procedure described in Section \ref{subsec:commonsense} is applied, whereby the retrieved paragraphs in Section \ref{subsec:commonsense} are replaced with the relevant knowledge from the current part.

\subsection{Evaluation}
\label{subsec:evaluation}

\paragraph{Experimental settings.} In all experiments, we utilize GPT-3 \texttt{text-davinci-002} unless otherwise stated. The maximum number of tokens for generation during completion is set to $256$. For zero-shot, few-shot, and chain-of-thought prompting, the temperature is fixed at $0$. For self-consistency and rethinking with retrieval, we randomly sample $10$ outputs\footnote{For commonsense reasoning, we sample $9$ outputs, as we have found that odd numbers of outputs tend to yield better voting performance for self-consistency on StrategyQA.} with temperature $0.7$. Detailed prompts can be found in Appendix \ref{subsec:prompts}. We evaluate the performance of different methods on commonsense and tabular reasoning using accuracy, and on temporal reasoning using the exact match metric as defined in \citet{rajpurkar2016squad}.

\paragraph{Results.} As shown in Table \ref{table:gpt3-results}, our proposed method, rethinking with retrieval, consistently outperforms all baselines on all three reasoning tasks without requiring additional training or fine-tuning. The results highlight the effectiveness of our approach in leveraging external knowledge to improve the performance of LLMs.

\begin{table*}[t]
\centering
\scalebox{1.0}{
\begin{tabular}{c|m{0.4\linewidth}|m{0.4\linewidth}}
\Xhline{2\arrayrulewidth}
  & \multicolumn{1}{c|}{Wrong supporting facts} & \multicolumn{1}{c}{Wrong inference} \bigstrut[t] \bigstrut[b] \\ \hline
Question & Was Lil Jon's top ranked Billboard song a collaboration with a member of The Lox?
 & Would the top of Mount Fuji stick out of the Sea of Japan? 
 \bigstrut[t] \bigstrut[b] \\ \hline
Answer & \multicolumn{1}{c|}{No} & \multicolumn{1}{c}{Yes} \bigstrut[t] \bigstrut[b]  \\ \hline
Facts & Lil Jon's highest ranked billboard song was Yeah; Yeah was a collaboration between Lil Jon, Usher, and Ludacris; The Lox is a rap trio consisting of: Styles P, Sheek Louch, and Jadakiss. & The average depth of the Sea of Japan is  5,748 feet (1,752 metres) and its maximum depth is 12,276 feet (3,742 metres); Mount Fuji is 3,776.24 metres (12,389.2 ft) tall.  \bigstrut[t] \bigstrut[b]  \\ \hline
GPT-3 & \textcolor{red}{Lil Jon's top ranked Billboard song was "Get Low"} with the East Side Boyz. The Lox is not a member of the East Side Boyz. Thus, Lil Jon's top ranked Billboard song was not a collaboration with a member of The Lox. So the answer is no.  & Mount Fuji is 3,776 meters tall. The Sea of Japan is about 3,741 meters deep. \textcolor{red}{Thus, the top of Mount Fuji would not stick out of the Sea of Japan.} So the answer is no.
 \bigstrut[t] \bigstrut[b]  \\ \Xhline{2\arrayrulewidth}
\end{tabular}}
\caption{Examples of incorrect outputs from GPT-3 with CoT prompting.} 
\label{table:gpt-3-outputs}
\end{table*}

\section{Analysis}

In this section, we perform a thorough analysis to gain a deeper understanding of \NAME{}.

\subsection{Limitations of LLMs in Reasoning}
\label{subsec:limitations}
In this subsection, we present an analysis of GPT-3 with CoT prompting on the StrategyQA dataset. Upon closer examination of the outputs of GPT-3, we observed that it can provide reasonable explanations and correct predictions for a number of questions. For example, when given the question ``\textit{Will the Albany in Georgia reach a hundred thousand occupants before the one in New York?}'', GPT-3 produced the following output:

\begin{quote}
\textcolor{cyan}{The Albany in New York has a population of about 98,000. The Albany in Georgia has a population of about 77,000.} \textcolor{YellowOrange}{Thus, the Albany in New York is more populous than the Albany in Georgia.} \textcolor{LimeGreen}{So the answer is no.}
\end{quote}

The above output consists of three components: (1) supporting facts (in cyan) that are based on a particular perspective, (2) chaining arguments (in orange), and (3) a prediction (in green). Components (1) and (2) contribute to the explanation. Overall, the output exhibits a high level of quality. However, we also observed that GPT-3 may occasionally produce incorrect supporting facts for its explanations or make incorrect inferences for its predictions, despite generally being able to identify suitable perspectives. 

\paragraph{Wrong supporting facts.} As shown in Table \ref{table:gpt-3-outputs}, GPT-3 provides the incorrect supporting fact for Lil Jon's top-ranked Billboard song, stating that it was ``Get Low'' instead of the correct answer, ``Yeah''. However, it does have the correct perspective on how to answer the question, ``\textit{Was Lil Jon’s top ranked Billboard song a collaboration with a member of
The Lox?}''.

\paragraph{Wrong inference.} As shown in Table \ref{table:gpt-3-outputs}, GPT-3 makes an incorrect inference, stating that the top of Mount Fuji ``would not stick out'' of the Sea of Japan, rather than the correct answer, ``would stick out''. However, it does provide correct supporting facts based on the appropriate perspective for the question, ``\textit{Would the top of Mount Fuji stick out of the Sea of Japan?}''. 

\subsection{Ablation Study}

\begin{table}[t]
\centering
\scalebox{0.92}{
\begin{tabular}{c|c|c}
\hline
Retrieval & Commonsense & Tabular  \bigstrut[t]  \\ \hline
Query-based & 73.36 & 36.69 \bigstrut[t] \bigstrut[b] \\ \hline
Decomposition-based & {\bf 77.73} & {\bf 39.05}  \bigstrut[t] \bigstrut[b] \\ \hline
\end{tabular}}
\caption{Comparison of query-based and decomposition-based retrieval on commonsense and tabular reasoning.
}
\label{table:retrieval-analysis}
\end{table} 

\paragraph{Importance of decomposition-based retrieval.} In our proposed method, we retrieve relevant external knowledge based on the decomposed reasoning steps rather than the original query. To further investigate the impact of this choice, we conducted additional experiments in which we used the original query for knowledge retrieval while keeping other aspects of our method unchanged. As shown in Table \ref{table:retrieval-analysis}, the results for these experiments are poor for both commonsense and temporal reasoning, indicating the importance of using decomposition-based retrieval in our approach.

\begin{table}[t]
\centering
\scalebox{1.0}{
\begin{tabular}{c|c}
\hline
Knowledge & Tabular  \bigstrut[t]  \\ \hline
External & 79.92 \bigstrut[t] \bigstrut[b] \\ \hline
Background &  84.75 \bigstrut[t] \bigstrut[b] \\ \hline
Background + External & {\bf 84.83}  \bigstrut[t] \bigstrut[b] \\ \hline
\end{tabular}}
\caption{Performance of \NAME{} with different types of knowledge on tabular reasoning: external only, background only, and a combination of both. External knowledge refers to WordNet and ConceptNet, while background knowledge refers to the tables.
}
\label{table:knowledge-analysis}
\end{table} 
\begin{figure*}[t]
		\centering
        \hspace{0.01in}
		\subfigure[Accuracy of predictions]{
			\centering
			\includegraphics[scale=0.35]{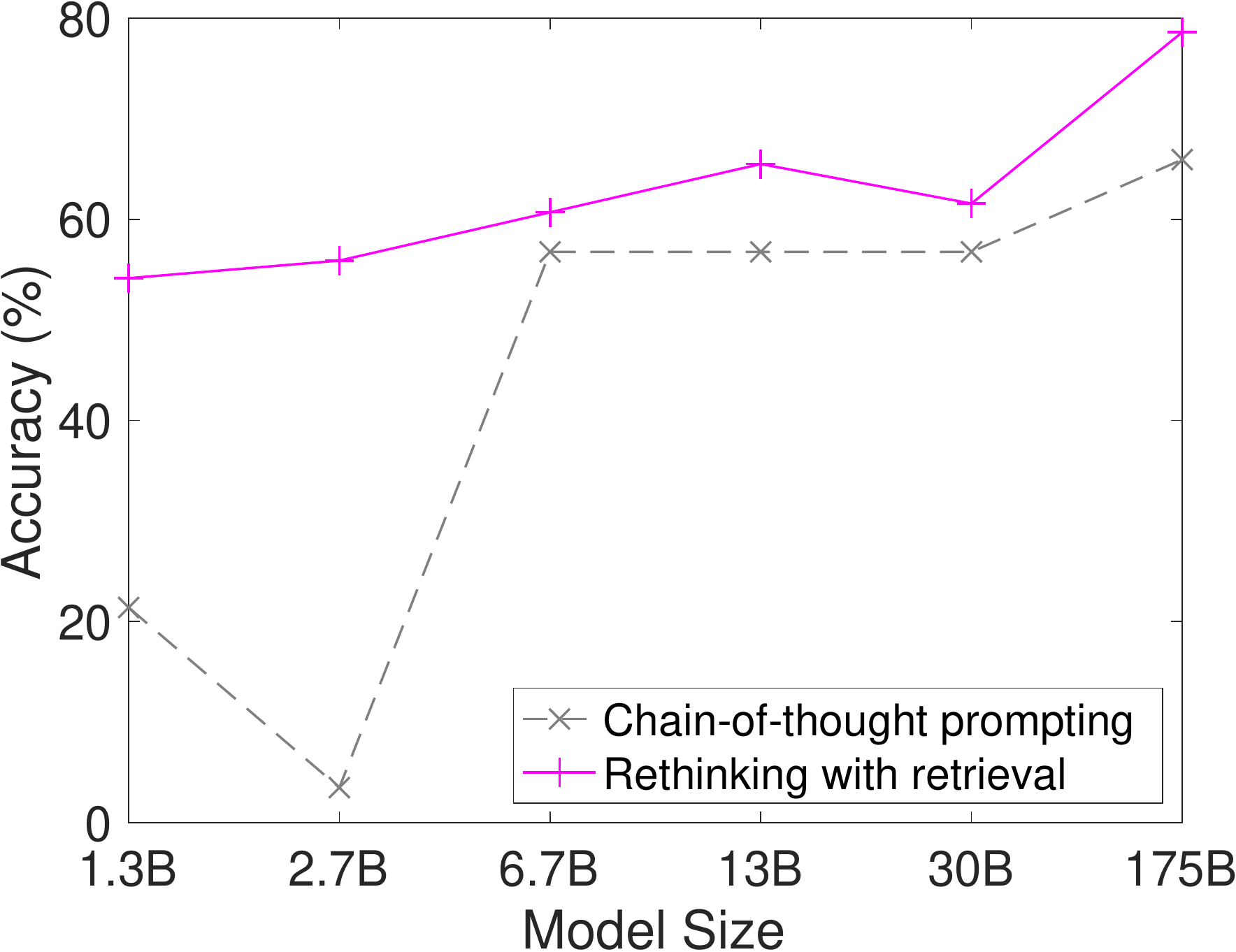}
			\label{fig:answer-accuracy-model-size}}
        \hspace{0.01in} 
        	\subfigure[Faithfulness of explanations]{
			\centering
		\includegraphics[scale=0.35]{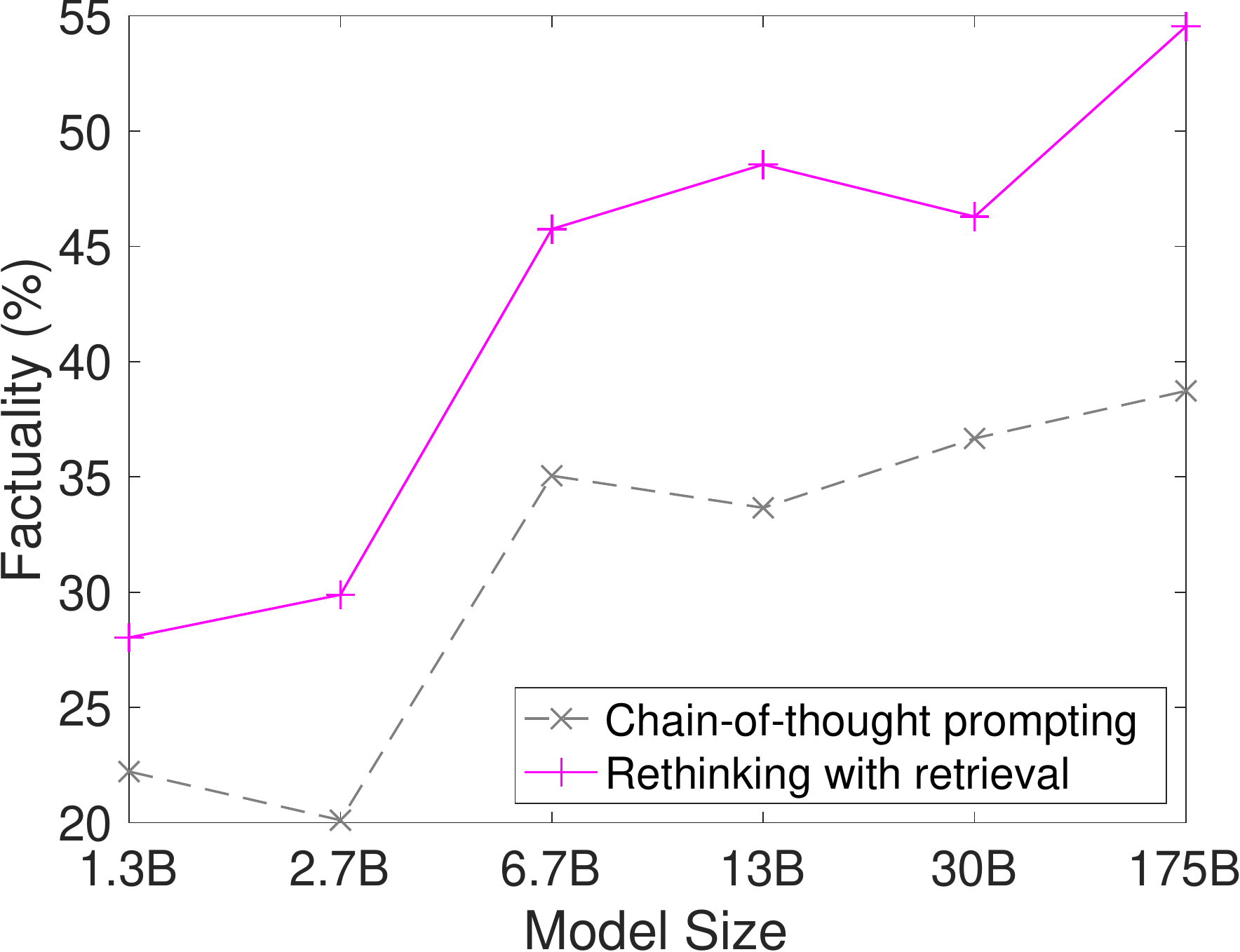}
			\label{fig:reason-factuality-model-size}}
		\caption{The effect of LM size on the performance of our proposed method (Variant II) and CoT prompting. We use various sizes of OPT models, with the exception of the 175B model, which is GPT-3.}
\label{fig:model-size}
\end{figure*}
\paragraph{The impact of different types of knowledge.} For tabular reasoning, we use both external knowledge (WordNet and ConceptNet) and background knowledge (tables) in our experiments. In this section, we further examine the effect of different types of knowledge on the performance of our proposed method. As shown in Table \ref{table:knowledge-analysis}, the additional improvement gained by incorporating Wikidata and ConceptNet in addition to tables is limited, indicating that GPT-3 already captures many word-level relations in these external knowledge sources. In addition, the observed significant improvement in tabular reasoning from using tables alone suggests that our proposed method can also effectively leverage background knowledge.

\subsection{Variations of the Proposed Approach}
\label{subsec:variations}
\paragraph{Basic approach: Weighting outputs.} In Section \ref{sec:framework}, we present a basic version of our proposal for taking advantage of external knowledge. Our basic approach involves \textit{weighting outputs as individual units} and using a \textit{voting} mechanism to select the best-supported prediction. We can also directly choose the best-supported output, which includes both an explanation and a prediction, without using voting. For example, in the running example of ``\textit{Did Aristotle use a laptop?}'' (see more in Section \ref{sec:framework}), the third reasoning path $R_3$ is the output most supported by the knowledge paragraphs $K_1$ and $K_2$.

\paragraph{Variant \uppercase\expandafter{\romannumeral1}: Fact selection.} The first variant of our approach involves selecting facts from the outputs of LLMs based on external knowledge. For example, consider the running example of ``\textit{Did Aristotle use a laptop?}'', where we only have access to the first two reasoning paths, $R_1$ and $R_2$. In this case, the first sentence in $R_2$ and the second sentence in $R_1$ are supported by knowledge $K_1$ and $K_2$, respectively. Therefore, the first variant would output the first sentence in $R_2$ and the second sentence in $R_1$ as the supporting facts.

\paragraph{Variant \uppercase\expandafter{\romannumeral2}: Fact generation.} The second variant of our approach involves generating facts based on both the outputs of LLMs and external knowledge. For example, consider the running example of ``\textit{Did Aristotle use a laptop?}'', where we only have access to the first reasoning path $R_1$. The second sentence in $R_1$ is supported by the second knowledge paragraph $K_2$. However, the first sentence is not supported by any evidence paragraphs. We can generate questions about the first sentence, such as ``When did Aristotle die?'' and use the first knowledge paragraph $K_1$ to generate a new fact: ``Aristotle died in 322BC.''. As a result, the second variant would output the generated fact ``Aristotle died in 322 BC.'' and the second sentence in $R_1$ as the supporting facts.

\paragraph{Inference with supporting facts.} For the two variants of our approach, we only have the supporting facts and need to perform a final inference step to obtain the corresponding prediction. One option for this inference is to use LLMs, but they can be costly \cite{brown2020language} or difficult to use \cite{zhang2022opt}. An alternative is to use an off-the-shelf model for inference with supporting facts, such as UnifiedQA \cite{khashabi2020unifiedqa, khashabi2022unifiedqa}. As discussed in Appendix \ref{subsec:inference-comparison}, UnifiedQA is more robust to noisy supporting facts than GPT-3. We thus use the second version of UnifiedQA, UnifiedQA-v2 \cite{khashabi2022unifiedqa}, for the final step of inference.

\paragraph{Experimental settings.} In this part, we focus on commonsense reasoning and use the \textit{evidence paragraphs} provided in StrategyQA as the relevant knowledge, rather than the retrieved paragraphs discussed in Section \ref{subsec:commonsense}. To evaluate the quality of the explanations, we adopt the best metric for factual consistency evaluation in \citet{honovich2022true}. For simplicity, we use the pre-trained NLI model released by \citet{nie2020adversarial} to compute the NLI-based metric, rather than fine-tuning T5-11B \cite{raffel2020exploring} ourselves. The implementation details of the two variants can be found in Appendix \ref{subsec:variants-implementation}.

\begin{table}[t]
\centering
\scalebox{0.88}{
\begin{tabular}{c|c|c}
\hline
Methods & Accuracy (\%) & Faithfulness (\%) \bigstrut[t]  \\ \hline
CoT prompting & 65.94 & 38.73 \bigstrut[t] \bigstrut[b] \\ \hline \hline
Basic (w/o voting) & 76.86 & 50.02 \bigstrut[t] \bigstrut[b] \\ \hline
Variant \uppercase\expandafter{\romannumeral1} & {\bf 78.60} & 54.11 \bigstrut[t] \bigstrut[b] \\ \hline
Variant \uppercase\expandafter{\romannumeral2} & {\bf 78.60} & {\bf 54.54} \bigstrut[t] \bigstrut[b] \\ \hline
\end{tabular}}
\caption{Comparison of various variations of \NAME{} and the CoT prompting baseline on StrategyQA using evidence paragraphs.
}
\label{table:proposal-variants}
\end{table}

\paragraph{Results.} Table \ref{table:proposal-variants} illustrates that the fact selection and fact generation variants of our proposal improve the faithfulness of the supporting facts in explanations, leading to increased prediction accuracy compared to the basic approach without voting. Across all variations of our proposal, we observe significant improvements in both prediction accuracy and the faithfulness of explanations when compared to the CoT prompting baseline.

The incorporation of a voting mechanism leads to an increased prediction accuracy of $79.91\%$ for the basic approach. Comparison with the performance (i.e., $77.73\%$) of the same approach using retrieved paragraphs rather than evidence paragraphs in Table \ref{table:gpt3-results} demonstrates that retrieved paragraphs are also effective for our proposal, as both significantly outperform the voting baseline, self-consistency (i.e., $73.36\%$), as shown in Table \ref{table:gpt3-results}.

It is noteworthy that UnifiedQA performs poorly on StrategyQA, achieving an accuracy of only $58.95\%$. However, when provided with gold supporting facts in StrategyQA, UnifiedQA demonstrates excellent performance with an accuracy of $90.83\%$. This suggests that UnifiedQA is suitable for last-step inference, but not effective for answering questions in StrategyQA.

\subsection{Impact of the Size of LMs}
\label{subsec:model-size}
In this subsection, we examine the effect of the size of LMs on the performance of our proposed method, specifically in the context of the fact generation variant. We compare the performance of our method using various sizes of OPT models \cite{zhang2022opt} in addition to GPT-3 (175B) using the same experimental setup as in Section \ref{subsec:variations}. As shown in Figure~\ref{fig:model-size}, our proposed method (Variant II) consistently outperforms CoT prompting in terms of both prediction accuracy and the faithfulness of explanations, even when using smaller LMs.

\section{Conclusion}

In conclusion, the proposed approach is a promising solution for utilizing external knowledge to assist LLMs. Unlike traditional methods, \NAME{} does not require additional training or fine-tuning, making it a lightweight and feasible option for LLMs. Through extensive experiments on three reasoning tasks using GPT-3, we have shown that \NAME{} is able to produce more faithful explanations and improve the performance of LLMs.  In the future, we plan to investigate various variations of \NAME{} to enhance its effectiveness and efficiency in augmenting LLMs with external knowledge. 


\clearpage
\bibliography{new}

\begin{thebibliography}{52}
\expandafter\ifx\csname natexlab\endcsname\relax\def\natexlab#1{#1}\fi

\bibitem[{Bollacker et~al.(2008)Bollacker, Evans, Paritosh, Sturge, and
  Taylor}]{bollacker2008freebase}
Kurt Bollacker, Colin Evans, Praveen Paritosh, Tim Sturge, and Jamie Taylor.
  2008.
\newblock Freebase: a collaboratively created graph database for structuring
  human knowledge.
\newblock In \emph{Proceedings of the 2008 ACM SIGMOD international conference
  on Management of data}, pages 1247--1250.

\bibitem[{Borgeaud et~al.(2021)Borgeaud, Mensch, Hoffmann, Cai, Rutherford,
  Millican, Driessche, Lespiau, Damoc, Clark et~al.}]{borgeaud2021improving}
Sebastian Borgeaud, Arthur Mensch, Jordan Hoffmann, Trevor Cai, Eliza
  Rutherford, Katie Millican, George van~den Driessche, Jean-Baptiste Lespiau,
  Bogdan Damoc, Aidan Clark, et~al. 2021.
\newblock Improving language models by retrieving from trillions of tokens.
\newblock \emph{arXiv preprint arXiv:2112.04426}.

\bibitem[{Brown et~al.(2020)Brown, Mann, Ryder, Subbiah, Kaplan, Dhariwal,
  Neelakantan, Shyam, Sastry, Askell et~al.}]{brown2020language}
Tom Brown, Benjamin Mann, Nick Ryder, Melanie Subbiah, Jared~D Kaplan, Prafulla
  Dhariwal, Arvind Neelakantan, Pranav Shyam, Girish Sastry, Amanda Askell,
  et~al. 2020.
\newblock Language models are few-shot learners.
\newblock \emph{Advances in neural information processing systems},
  33:1877--1901.

\bibitem[{Chowdhery et~al.(2022)Chowdhery, Narang, Devlin, Bosma, Mishra,
  Roberts, Barham, Chung, Sutton, Gehrmann et~al.}]{chowdhery2022palm}
Aakanksha Chowdhery, Sharan Narang, Jacob Devlin, Maarten Bosma, Gaurav Mishra,
  Adam Roberts, Paul Barham, Hyung~Won Chung, Charles Sutton, Sebastian
  Gehrmann, et~al. 2022.
\newblock Palm: Scaling language modeling with pathways.
\newblock \emph{arXiv preprint arXiv:2204.02311}.

\bibitem[{Cobbe et~al.(2021)Cobbe, Kosaraju, Bavarian, Hilton, Nakano, Hesse,
  and Schulman}]{cobbe2021training}
Karl Cobbe, Vineet Kosaraju, Mohammad Bavarian, Jacob Hilton, Reiichiro Nakano,
  Christopher Hesse, and John Schulman. 2021.
\newblock Training verifiers to solve math word problems.
\newblock \emph{arXiv preprint arXiv:2110.14168}.

\bibitem[{Dagan et~al.(2005)Dagan, Glickman, and Magnini}]{dagan2005pascal}
Ido Dagan, Oren Glickman, and Bernardo Magnini. 2005.
\newblock The pascal recognising textual entailment challenge.
\newblock In \emph{Machine learning challenges workshop}, pages 177--190.
  Springer.

\bibitem[{Deutsch et~al.(2021)Deutsch, Bedrax-Weiss, and
  Roth}]{deutsch2021towards}
Daniel Deutsch, Tania Bedrax-Weiss, and Dan Roth. 2021.
\newblock Towards question-answering as an automatic metric for evaluating the
  content quality of a summary.
\newblock \emph{Transactions of the Association for Computational Linguistics},
  9:774--789.

\bibitem[{Devlin et~al.(2019)Devlin, Chang, Lee, and
  Toutanova}]{devlin2019bert}
Jacob Devlin, Ming-Wei Chang, Kenton Lee, and Kristina Toutanova. 2019.
\newblock {BERT}: Pre-training of deep bidirectional transformers for language
  understanding.
\newblock In \emph{Proceedings of the 2019 Conference of the North American
  Chapter of the Association for Computational Linguistics: Human Language
  Technologies, Volume 1 (Long and Short Papers)}, pages 4171--4186.

\bibitem[{Fabbri et~al.(2021)Fabbri, Wu, Liu, and Xiong}]{fabbri2021qafacteval}
Alexander~R Fabbri, Chien-Sheng Wu, Wenhao Liu, and Caiming Xiong. 2021.
\newblock Qafacteval: Improved qa-based factual consistency evaluation for
  summarization.
\newblock \emph{arXiv preprint arXiv:2112.08542}.

\bibitem[{Geva et~al.(2021)Geva, Khashabi, Segal, Khot, Roth, and
  Berant}]{geva2021did}
Mor Geva, Daniel Khashabi, Elad Segal, Tushar Khot, Dan Roth, and Jonathan
  Berant. 2021.
\newblock Did aristotle use a laptop? a question answering benchmark with
  implicit reasoning strategies.
\newblock \emph{Transactions of the Association for Computational Linguistics},
  9:346--361.

\bibitem[{Gui et~al.(2021)Gui, Wang, Huang, Hauptmann, Bisk, and
  Gao}]{gui2021kat}
Liangke Gui, Borui Wang, Qiuyuan Huang, Alex Hauptmann, Yonatan Bisk, and
  Jianfeng Gao. 2021.
\newblock Kat: A knowledge augmented transformer for vision-and-language.
\newblock \emph{arXiv preprint arXiv:2112.08614}.

\bibitem[{Gupta et~al.(2020)Gupta, Mehta, Nokhiz, and
  Srikumar}]{gupta2020infotabs}
Vivek Gupta, Maitrey Mehta, Pegah Nokhiz, and Vivek Srikumar. 2020.
\newblock Infotabs: Inference on tables as semi-structured data.
\newblock In \emph{Proceedings of the 58th Annual Meeting of the Association
  for Computational Linguistics}, pages 2309--2324.

\bibitem[{Guu et~al.(2020)Guu, Lee, Tung, Pasupat, and
  Chang}]{guu2020retrieval}
Kelvin Guu, Kenton Lee, Zora Tung, Panupong Pasupat, and Mingwei Chang. 2020.
\newblock Retrieval augmented language model pre-training.
\newblock In \emph{International Conference on Machine Learning}, pages
  3929--3938. PMLR.

\bibitem[{Honovich et~al.(2022)Honovich, Aharoni, Herzig, Taitelbaum,
  Kukliansy, Cohen, Scialom, Szpektor, Hassidim, and Matias}]{honovich2022true}
Or~Honovich, Roee Aharoni, Jonathan Herzig, Hagai Taitelbaum, Doron Kukliansy,
  Vered Cohen, Thomas Scialom, Idan Szpektor, Avinatan Hassidim, and Yossi
  Matias. 2022.
\newblock True: Re-evaluating factual consistency evaluation.
\newblock In \emph{Proceedings of the Second DialDoc Workshop on
  Document-grounded Dialogue and Conversational Question Answering}, pages
  161--175.

\bibitem[{Honovich et~al.(2021)Honovich, Choshen, Aharoni, Neeman, Szpektor,
  and Abend}]{honovich2021q2}
Or~Honovich, Leshem Choshen, Roee Aharoni, Ella Neeman, Idan Szpektor, and Omri
  Abend. 2021.
\newblock Q2:: Evaluating factual consistency in knowledge-grounded dialogues
  via question generation and question answering.
\newblock In \emph{Proceedings of the 2021 Conference on Empirical Methods in
  Natural Language Processing}, pages 7856--7870.

\bibitem[{Jia et~al.(2018)Jia, Abujabal, Saha~Roy, Str{\"o}tgen, and
  Weikum}]{jia2018tempquestions}
Zhen Jia, Abdalghani Abujabal, Rishiraj Saha~Roy, Jannik Str{\"o}tgen, and
  Gerhard Weikum. 2018.
\newblock Tempquestions: A benchmark for temporal question answering.
\newblock In \emph{Companion Proceedings of the The Web Conference 2018}, pages
  1057--1062.

\bibitem[{Joshi et~al.(2020)Joshi, Lee, Luan, and
  Toutanova}]{joshi2020contextualized}
Mandar Joshi, Kenton Lee, Yi~Luan, and Kristina Toutanova. 2020.
\newblock Contextualized representations using textual encyclopedic knowledge.
\newblock \emph{arXiv preprint arXiv:2004.12006}.

\bibitem[{Karpukhin et~al.(2020)Karpukhin, Oguz, Min, Lewis, Wu, Edunov, Chen,
  and Yih}]{karpukhin2020dense}
Vladimir Karpukhin, Barlas Oguz, Sewon Min, Patrick Lewis, Ledell Wu, Sergey
  Edunov, Danqi Chen, and Wen-tau Yih. 2020.
\newblock Dense passage retrieval for open-domain question answering.
\newblock In \emph{Proceedings of the 2020 Conference on Empirical Methods in
  Natural Language Processing (EMNLP)}, pages 6769--6781.

\bibitem[{Khandelwal et~al.(2020)Khandelwal, Levy, Jurafsky, Zettlemoyer, and
  Lewis}]{khandelwal2019generalization}
Urvashi Khandelwal, Omer Levy, Dan Jurafsky, Luke Zettlemoyer, and Mike Lewis.
  2020.
\newblock Generalization through memorization: Nearest neighbor language
  models.
\newblock In \emph{International Conference on Learning Representations}.

\bibitem[{Khashabi et~al.(2022)Khashabi, Kordi, and
  Hajishirzi}]{khashabi2022unifiedqa}
Daniel Khashabi, Yeganeh Kordi, and Hannaneh Hajishirzi. 2022.
\newblock Unifiedqa-v2: Stronger generalization via broader cross-format
  training.
\newblock \emph{arXiv preprint arXiv:2202.12359}.

\bibitem[{Khashabi et~al.(2020)Khashabi, Min, Khot, Sabharwal, Tafjord, Clark,
  and Hajishirzi}]{khashabi2020unifiedqa}
Daniel Khashabi, Sewon Min, Tushar Khot, Ashish Sabharwal, Oyvind Tafjord,
  Peter Clark, and Hannaneh Hajishirzi. 2020.
\newblock Unifiedqa: Crossing format boundaries with a single qa system.
\newblock In \emph{Findings of the Association for Computational Linguistics:
  EMNLP 2020}, pages 1896--1907.

\bibitem[{Kojima et~al.(2022)Kojima, Gu, Reid, Matsuo, and
  Iwasawa}]{kojima2022large}
Takeshi Kojima, Shixiang~Shane Gu, Machel Reid, Yutaka Matsuo, and Yusuke
  Iwasawa. 2022.
\newblock Large language models are zero-shot reasoners.
\newblock \emph{arXiv preprint arXiv:2205.11916}.

\bibitem[{Komeili et~al.(2022)Komeili, Shuster, and
  Weston}]{komeili2022internet}
Mojtaba Komeili, Kurt Shuster, and Jason Weston. 2022.
\newblock Internet-augmented dialogue generation.
\newblock In \emph{Proceedings of the 60th Annual Meeting of the Association
  for Computational Linguistics (Volume 1: Long Papers)}, pages 8460--8478.

\bibitem[{Lewis et~al.(2020)Lewis, Perez, Piktus, Petroni, Karpukhin, Goyal,
  K{\"u}ttler, Lewis, Yih, Rockt{\"a}schel et~al.}]{lewis2020retrieval}
Patrick Lewis, Ethan Perez, Aleksandra Piktus, Fabio Petroni, Vladimir
  Karpukhin, Naman Goyal, Heinrich K{\"u}ttler, Mike Lewis, Wen-tau Yih, Tim
  Rockt{\"a}schel, et~al. 2020.
\newblock Retrieval-augmented generation for knowledge-intensive nlp tasks.
\newblock \emph{Advances in Neural Information Processing Systems},
  33:9459--9474.

\bibitem[{Lin et~al.(2021)Lin, Ma, Lin, Yang, Pradeep, and
  Nogueira}]{Lin_etal_SIGIR2021_Pyserini}
Jimmy Lin, Xueguang Ma, Sheng-Chieh Lin, Jheng-Hong Yang, Ronak Pradeep, and
  Rodrigo Nogueira. 2021.
\newblock {Pyserini}: A {Python} toolkit for reproducible information retrieval
  research with sparse and dense representations.
\newblock In \emph{Proceedings of the 44th Annual International ACM SIGIR
  Conference on Research and Development in Information Retrieval (SIGIR
  2021)}, pages 2356--2362.

\bibitem[{Liu et~al.(2022)Liu, Liu, Lu, Welleck, West, Le~Bras, Choi, and
  Hajishirzi}]{liu2022generated}
Jiacheng Liu, Alisa Liu, Ximing Lu, Sean Welleck, Peter West, Ronan Le~Bras,
  Yejin Choi, and Hannaneh Hajishirzi. 2022.
\newblock Generated knowledge prompting for commonsense reasoning.
\newblock In \emph{Proceedings of the 60th Annual Meeting of the Association
  for Computational Linguistics (Volume 1: Long Papers)}, pages 3154--3169.

\bibitem[{Liu et~al.(2019)Liu, Ott, Goyal, Du, Joshi, Chen, Levy, Lewis,
  Zettlemoyer, and Stoyanov}]{liu2019roberta}
Yinhan Liu, Myle Ott, Naman Goyal, Jingfei Du, Mandar Joshi, Danqi Chen, Omer
  Levy, Mike Lewis, Luke Zettlemoyer, and Veselin Stoyanov. 2019.
\newblock Roberta: A robustly optimized bert pretraining approach.
\newblock \emph{arXiv preprint arXiv:1907.11692}.

\bibitem[{Miller(1995)}]{miller1995wordnet}
George~A Miller. 1995.
\newblock Wordnet: a lexical database for english.
\newblock \emph{Communications of the ACM}, 38(11):39--41.

\bibitem[{Nakano et~al.(2021)Nakano, Hilton, Balaji, Wu, Ouyang, Kim, Hesse,
  Jain, Kosaraju, Saunders et~al.}]{nakano2021webgpt}
Reiichiro Nakano, Jacob Hilton, Suchir Balaji, Jeff Wu, Long Ouyang, Christina
  Kim, Christopher Hesse, Shantanu Jain, Vineet Kosaraju, William Saunders,
  et~al. 2021.
\newblock Webgpt: Browser-assisted question-answering with human feedback.
\newblock \emph{arXiv preprint arXiv:2112.09332}.

\bibitem[{Neeraja et~al.(2021)Neeraja, Gupta, and
  Srikumar}]{neeraja2021incorporating}
J~Neeraja, Vivek Gupta, and Vivek Srikumar. 2021.
\newblock Incorporating external knowledge to enhance tabular reasoning.
\newblock In \emph{Proceedings of the 2021 Conference of the North American
  Chapter of the Association for Computational Linguistics: Human Language
  Technologies}, pages 2799--2809.

\bibitem[{Nie et~al.(2020)Nie, Williams, Dinan, Bansal, Weston, and
  Kiela}]{nie2020adversarial}
Yixin Nie, Adina Williams, Emily Dinan, Mohit Bansal, Jason Weston, and Douwe
  Kiela. 2020.
\newblock Adversarial nli: A new benchmark for natural language understanding.
\newblock In \emph{Proceedings of the 58th Annual Meeting of the Association
  for Computational Linguistics}, pages 4885--4901.

\bibitem[{Nye et~al.(2022)Nye, Andreassen, Gur-Ari, Michalewski, Austin,
  Bieber, Dohan, Lewkowycz, Bosma, Luan et~al.}]{nye2022show}
Maxwell Nye, Anders~Johan Andreassen, Guy Gur-Ari, Henryk Michalewski, Jacob
  Austin, David Bieber, David Dohan, Aitor Lewkowycz, Maarten Bosma, David
  Luan, et~al. 2022.
\newblock Show your work: Scratchpads for intermediate computation with
  language models.
\newblock In \emph{Deep Learning for Code Workshop}.

\bibitem[{Nye et~al.(2021)Nye, Tessler, Tenenbaum, and Lake}]{nye2021improving}
Maxwell Nye, Michael Tessler, Josh Tenenbaum, and Brenden~M Lake. 2021.
\newblock Improving coherence and consistency in neural sequence models with
  dual-system, neuro-symbolic reasoning.
\newblock \emph{Advances in Neural Information Processing Systems},
  34:25192--25204.

\bibitem[{Ouyang et~al.(2022)Ouyang, Wu, Jiang, Almeida, Wainwright, Mishkin,
  Zhang, Agarwal, Slama, Ray et~al.}]{ouyang2022training}
Long Ouyang, Jeff Wu, Xu~Jiang, Diogo Almeida, Carroll~L Wainwright, Pamela
  Mishkin, Chong Zhang, Sandhini Agarwal, Katarina Slama, Alex Ray, et~al.
  2022.
\newblock Training language models to follow instructions with human feedback.
\newblock \emph{arXiv preprint arXiv:2203.02155}.

\bibitem[{Raffel et~al.(2020)Raffel, Shazeer, Roberts, Lee, Narang, Matena,
  Zhou, Li, and Liu}]{raffel2020exploring}
Colin Raffel, Noam Shazeer, Adam Roberts, Katherine Lee, Sharan Narang, Michael
  Matena, Yanqi Zhou, Wei Li, and Peter~J Liu. 2020.
\newblock Exploring the limits of transfer learning with a unified text-to-text
  transformer.
\newblock \emph{Journal of Machine Learning Research}, 21:1--67.

\bibitem[{Rajpurkar et~al.(2016)Rajpurkar, Zhang, Lopyrev, and
  Liang}]{rajpurkar2016squad}
Pranav Rajpurkar, Jian Zhang, Konstantin Lopyrev, and Percy Liang. 2016.
\newblock Squad: 100,000+ questions for machine comprehension of text.
\newblock In \emph{Proceedings of the 2016 Conference on Empirical Methods in
  Natural Language Processing}, pages 2383--2392.

\bibitem[{Robertson et~al.(2009)Robertson, Zaragoza
  et~al.}]{robertson2009probabilistic}
Stephen Robertson, Hugo Zaragoza, et~al. 2009.
\newblock The probabilistic relevance framework: Bm25 and beyond.
\newblock \emph{Foundations and Trends{\textregistered} in Information
  Retrieval}, 3(4):333--389.

\bibitem[{Shuster et~al.(2022)Shuster, Komeili, Adolphs, Roller, Szlam, and
  Weston}]{shuster2022language}
Kurt Shuster, Mojtaba Komeili, Leonard Adolphs, Stephen Roller, Arthur Szlam,
  and Jason Weston. 2022.
\newblock Language models that seek for knowledge: Modular search \& generation
  for dialogue and prompt completion.
\newblock \emph{arXiv preprint arXiv:2203.13224}.

\bibitem[{Song et~al.(2020)Song, Tan, Qin, Lu, and Liu}]{song2020mpnet}
Kaitao Song, Xu~Tan, Tao Qin, Jianfeng Lu, and Tie-Yan Liu. 2020.
\newblock Mpnet: Masked and permuted pre-training for language understanding.
\newblock \emph{Advances in Neural Information Processing Systems},
  33:16857--16867.

\bibitem[{Speer et~al.(2017)Speer, Chin, and Havasi}]{speer2017conceptnet}
Robyn Speer, Joshua Chin, and Catherine Havasi. 2017.
\newblock Conceptnet 5.5: An open multilingual graph of general knowledge.
\newblock In \emph{Thirty-first AAAI conference on artificial intelligence}.

\bibitem[{Talmor et~al.(2020)Talmor, Tafjord, Clark, Goldberg, and
  Berant}]{talmor2020leap}
Alon Talmor, Oyvind Tafjord, Peter Clark, Yoav Goldberg, and Jonathan Berant.
  2020.
\newblock Leap-of-thought: Teaching pre-trained models to systematically reason
  over implicit knowledge.
\newblock \emph{Advances in Neural Information Processing Systems},
  33:20227--20237.

\bibitem[{Thoppilan et~al.(2022)Thoppilan, De~Freitas, Hall, Shazeer,
  Kulshreshtha, Cheng, Jin, Bos, Baker, Du et~al.}]{thoppilan2022lamda}
Romal Thoppilan, Daniel De~Freitas, Jamie Hall, Noam Shazeer, Apoorv
  Kulshreshtha, Heng-Tze Cheng, Alicia Jin, Taylor Bos, Leslie Baker, Yu~Du,
  et~al. 2022.
\newblock Lamda: Language models for dialog applications.
\newblock \emph{arXiv preprint arXiv:2201.08239}.

\bibitem[{Varun et~al.(2022)Varun, Sharma, and Gupta}]{varun2022trans}
Yerram Varun, Aayush Sharma, and Vivek Gupta. 2022.
\newblock Trans-kblstm: An external knowledge enhanced transformer bilstm model
  for tabular reasoning.
\newblock In \emph{Proceedings of Deep Learning Inside Out (DeeLIO 2022): The
  3rd Workshop on Knowledge Extraction and Integration for Deep Learning
  Architectures}, pages 62--78.

\bibitem[{Vaswani et~al.(2017)Vaswani, Shazeer, Parmar, Uszkoreit, Jones,
  Gomez, Kaiser, and Polosukhin}]{vaswani2017attention}
Ashish Vaswani, Noam Shazeer, Niki Parmar, Jakob Uszkoreit, Llion Jones,
  Aidan~N Gomez, {\L}ukasz Kaiser, and Illia Polosukhin. 2017.
\newblock Attention is all you need.
\newblock \emph{Advances in neural information processing systems}, 30.

\bibitem[{Vrande{\v{c}}i{\'c} and Kr{\"o}tzsch(2014)}]{vrandevcic2014wikidata}
Denny Vrande{\v{c}}i{\'c} and Markus Kr{\"o}tzsch. 2014.
\newblock Wikidata: a free collaborative knowledgebase.
\newblock \emph{Communications of the ACM}, 57(10):78--85.

\bibitem[{Wang et~al.(2022)Wang, Wei, Schuurmans, Le, Chi, and
  Zhou}]{wang2022self}
Xuezhi Wang, Jason Wei, Dale Schuurmans, Quoc Le, Ed~Chi, and Denny Zhou. 2022.
\newblock Self-consistency improves chain of thought reasoning in language
  models.
\newblock \emph{arXiv preprint arXiv:2203.11171}.

\bibitem[{Wei et~al.(2022)Wei, Wang, Schuurmans, Bosma, Chi, Le, and
  Zhou}]{wei2022chain}
Jason Wei, Xuezhi Wang, Dale Schuurmans, Maarten Bosma, Ed~Chi, Quoc Le, and
  Denny Zhou. 2022.
\newblock Chain of thought prompting elicits reasoning in large language
  models.
\newblock \emph{arXiv preprint arXiv:2201.11903}.

\bibitem[{Wolf et~al.(2020)Wolf, Debut, Sanh, Chaumond, Delangue, Moi, Cistac,
  Rault, Louf, Funtowicz et~al.}]{wolf2020transformers}
Thomas Wolf, Lysandre Debut, Victor Sanh, Julien Chaumond, Clement Delangue,
  Anthony Moi, Pierric Cistac, Tim Rault, R{\'e}mi Louf, Morgan Funtowicz,
  et~al. 2020.
\newblock Transformers: State-of-the-art natural language processing.
\newblock In \emph{Proceedings of the 2020 conference on empirical methods in
  natural language processing: system demonstrations}, pages 38--45.

\bibitem[{Ye and Durrett(2022)}]{ye2022unreliability}
Xi~Ye and Greg Durrett. 2022.
\newblock The unreliability of explanations in few-shot in-context learning.
\newblock \emph{arXiv preprint arXiv:2205.03401}.

\bibitem[{Zelikman et~al.(2022)Zelikman, Wu, and Goodman}]{zelikman2022star}
Eric Zelikman, Yuhuai Wu, and Noah~D Goodman. 2022.
\newblock Star: Bootstrapping reasoning with reasoning.
\newblock \emph{arXiv preprint arXiv:2203.14465}.

\bibitem[{Zhang et~al.(2022)Zhang, Roller, Goyal, Artetxe, Chen, Chen, Dewan,
  Diab, Li, Lin et~al.}]{zhang2022opt}
Susan Zhang, Stephen Roller, Naman Goyal, Mikel Artetxe, Moya Chen, Shuohui
  Chen, Christopher Dewan, Mona Diab, Xian Li, Xi~Victoria Lin, et~al. 2022.
\newblock Opt: Open pre-trained transformer language models.
\newblock \emph{arXiv preprint arXiv:2205.01068}.

\bibitem[{Zhou et~al.(2022)Zhou, Sch{\"a}rli, Hou, Wei, Scales, Wang,
  Schuurmans, Bousquet, Le, and Chi}]{zhou2022least}
Denny Zhou, Nathanael Sch{\"a}rli, Le~Hou, Jason Wei, Nathan Scales, Xuezhi
  Wang, Dale Schuurmans, Olivier Bousquet, Quoc Le, and Ed~Chi. 2022.
\newblock Least-to-most prompting enables complex reasoning in large language
  models.
\newblock \emph{arXiv preprint arXiv:2205.10625}.

\end{thebibliography}
\bibliographystyle{acl_natbib}

\clearpage
\appendix
\begin{table*}[ht]
\centering
\scalebox{1.0}{
\begin{tabular}{m{1.0\linewidth}}
\hline
{\bf Q:} who was governor of minnesota when maathaad maathaadu mallige was released? \bigstrut[t] \\
{\bf A:} Maathaad Maathaadu Mallige was released on 24 August 2007. Tim Pawlenty served as the 39th governor of Minnesota from 2003 to 2011. Thus, Tim Pawlenty was governor of minnesota when maathaad maathaadu mallige was released. {\bf So the answer is Tim Pawlenty.} \\
{\bf Q:} who was us president during the costa rican civil war? \bigstrut[t]  \\
{\bf A:} The Costa Rican civil war was a civil war in Costa Rica from 12 March to 24 April 1948. Harry S. Truman was the 33rd president of the United States, serving from 1945 to 1953. Thus, Harry S. Truman was us president during the costa rican civil war. {\bf So the answer is Harry S. Truman.} \\
{\bf Q:} who was governor of oregon when the collector was released? \bigstrut[t]\\
{\bf A:} The Collector premiered at the Cannes Film Festival on May 20, 1965. Mark Hatfield served as the 29th governor of Oregon from 1959 to 1967. Thus, Mark Hatfield was governor of oregon when the collector was released. {\bf So the answer is Mark Hatfield.}\\
{\bf Q:} who was governor of oregon when shanghai noon was released? \bigstrut[t] \\
{\bf A:} Shanghai Noon was released on May 26, 2000. John Kitzhaber served as the 35th governor of Oregon from 1995 to 2003. Thus, John Kitzhaber was governor of oregon when shanghai noon was released. {\bf So the answer is John Kitzhaber.}\\
{\bf Q:} who was us president when john andrew shulze was a teenager? \bigstrut[t] \\
{\bf A:} John Andrew Shulze was born on July 19, 1775. A teenager is someone who is between 13 and 19 years old. George Washington served as the first president of the United States from 1789 to 1797. Thus, George Washington was us president when john andrew shulze was a teenager. {\bf So the answer is George Washington.}\\
{\bf Q:} who was us president during the seventh coalition? \bigstrut[t] \\
{\bf A:} The War of the Seventh Coalition was from 20 March to 8 July 1815. James Madison served as the fourth president of the United States from 1809 to 1817. Thus, James Madison was us president during the seventh coalition. {\bf So the answer is James Madison.} \bigstrut[b]\\
\hline
\end{tabular}}
\caption{The CoT prompt for temporal reasoning.
}
\label{table:temporal-prompt}
\end{table*} 

\section{Appendix}
\label{sec:appendix}

In this section, we provide additional details on our experimental setup. Further information can be found in our code.

\subsection{Detailed Prompts}
\label{subsec:prompts}

We adopt the same CoT prompt for commonsense reasoning (i.e., StrategyQA) as those presented in \citet{wei2022chain}. The CoT prompt for temporal reasoning is provided in Table \ref{table:temporal-prompt}. For tabular reasoning, we adopt the method of \citet{brown2020language} for converting NLI into QA for RTE \cite{dagan2005pascal}, and randomly sample $6$ examples from the training data to construct the prompt, as shown in Table \ref{table:tabular-prompt}. The few-shot prompt utilizes the same exemplars as the CoT prompt and does not involve CoT reasoning processes.

\subsection{Description of Faithfulness Functions}
\label{subsec:faithfulness-functions}
For a sentence $s$, we denote its MPNet similarity, entailment score, and contradiction score as $M(s)$, $E(s)$, and $C(s)$, respectively. In our experiments, the corresponding thresholds for these scores are $T_m = 0.5$, $T_e = 0.6$, and $T_c = 0.99$. Given the entailment scores, contradiction scores, and MPNet similarities of all supporting facts (denoted as $S$) in the explanation of a reasoning path $R$, different faithfulness functions $f_{\mathcal{KB}}(\cdot)$ can be adopted in different settings as follows:
\begin{itemize}
    \item[(1)] $f_{\mathcal{KB}}(R) = \sum_{s \in S}
 [M(s) \times (M(s) >= T_m) + E(s) \times (M(s) < T_m)  - C(s)]$
    \item[(2)] $f_{\mathcal{KB}}(R) = \sum_{s \in S} [M(s) + E(s)]$
    \item[(3)] $f_{\mathcal{KB}}(R) = \sum_{s \in S} [E(s) \times (E(s) >= T_e) - C(s) \times (C(s) >= T_c)]$
\end{itemize}

In Section \ref{sec:experiments}, we employ function (1) for commonsense and tabular reasoning. For temporal reasoning, we use function (2) as the distinct nature of sentences converted from temporal relations leads to unreliable contradiction scores. In Sections \ref{subsec:variations}-\ref{subsec:model-size}, we use function (3) for commonsense reasoning with evidence paragraphs, as the high quality of the relevant knowledge negates the need for the complementary use of the MPNet similarity to improve the entailment score.

\subsection{Comparison of Retrieval Systems}
\label{subsec:retrieval-comparison}

For commonsense reasoning, we utilized different retrieval systems in \citet{karpukhin2020dense} to retrieve relevant paragraphs from Wikipedia. The performance of BM25, DPR, and BM25+DPR were $77.73\%$, $58.52\%$, and $77.29\%$, respectively, indicating that BM25 is the best choice in our case.

\subsection{Implementation Details for the Two Variants of \NAME{}}
\label{subsec:variants-implementation}

\paragraph{Fact selection implementation details.} In this work, we utilize the information present in the top-ranked output produced by our basic approach as a guide. To this end, we apply a greedy clustering algorithm to group the sentences from all outputs into distinct topic categories based on the cosine similarity of their MPNet sentence embeddings. For each fact in the top-ranked output of our basic approach, we identify the fact with the highest faithfulness within the same topic group and replace it in the output. The faithfulness of a fact is calculated using the $f_{\mathcal{KB}}$ function by replacing the supporting facts with a single fact.

\paragraph{Fact generation implementation details.} In this part, we generate questions for the named entities present in each fact of the top-ranked output produced by our basic approach, and retrieve the corresponding answers from the evidence paragraphs using UnifiedQA. We employ the question generation model described in \citet{deutsch2021towards}, which has been shown to be more extractive compared to other models as demonstrated in \citet{fabbri2021qafacteval}. We adopt the question filtering approach proposed in \citet{honovich2021q2} using an off-the-shelf extractive QA model (ktrapeznikov/albert-xlarge-v2-squad-v2 from Hugging Face \cite{wolf2020transformers}). We then use an off-the-shelf model (MarkS/bart-base-qa2d from Hugging Face) to convert the generated QA pairs into declarative sentences. We apply simple rules based on the entailment and contradiction scores of the selected facts from the fact selection variant and the generated declarative sentences to obtain the final generated facts.

\subsection{Comparison of Different Inference Methods with Supporting Facts}
\label{subsec:inference-comparison}

In our experiments, we utilize UnifiedQA for the final step of inference in both variants. However, it is worth noting that GPT-3 could also be used for this purpose. As shown in Table \ref{table:inference-comparison}, we observe that UnifiedQA performs better at inference with generated facts, while GPT-3 with CoT prompting performs better with empty or gold facts. This suggests that UnifiedQA is more robust to noisy inputs compared to GPT-3. Additionally, both UnifiedQA and GPT-3 with CoT prompting significantly outperform GPT-3 with zero-shot prompting, indicating that the CoT prompting is also beneficial for the final step of inference.

\begin{table}
\centering
\scalebox{0.85}{
\begin{tabular}{c|c|c}
 & Methods & Accuracy (\%) \bigstrut[b] \\ \hline
\multirow{3}{*}{Empty facts} &  GPT-3 (zero-shot) & 58.08 \bigstrut[t] \\ 
 & GPT-3 (CoT) & {\bf 65.94} \\
 & UnifiedQA & 58.95 \bigstrut[b] \\ \hline
 \multirow{3}{*}{Gold facts} & GPT-3 (zero-shot) & 81.66 \bigstrut[t] \\ 
 & GPT-3 (CoT) & {\bf 91.70} \\
 & UnifiedQA & 90.83 \bigstrut[b] \\ \hline
 \multirow{3}{*}{Generated facts} & GPT-3 (zero-shot) & 69.87 \bigstrut[t] \\ 
 & GPT-3 (CoT) & 76.42 \\
 & UnifiedQA & {\bf 78.60} \bigstrut[b]  \\ \Xhline{2\arrayrulewidth}
\end{tabular}}
\caption{Comparison of different inference methods on empty, gold, and generated facts.
}
\label{table:inference-comparison}
\end{table} 

\begin{table*}
\centering
\scalebox{1.0}{
\begin{tabular}{m{1.0\linewidth}}
\hline
Charles Sumner Tainter was Born on April 25, 1854   ( 1854-04-25 )   Watertown, Massachusetts, U.S..  Charles Sumner Tainter was Died on April 20, 1940   ( 1940-04-21 )  (aged 85)  San Diego, California, U.S..  The Nationality of Charles Sumner Tainter are American.  The Known for of Charles Sumner Tainter are Photophone, phonograph Father Of The Speaking Machine. \bigstrut[t] \\
{\bf Question:} Charles Sumner Tainter never left the state of Massachusetts. True or False?\\
{\bf Answer:} Charles Sumner Tainter was died in San Diego, California, U.S.. California is a state. Thus, Charles Sumner Tainter has left the state of Massachusetts. {\bf So the answer is false.} \\
The Region of Curitiba are South.  The Elevation of Curitiba are 934.6 m (3,066.3 ft).  The Density of Curitiba are 4,062/km 2  (10,523/sq mi).  The Metro density of Curitiba are 210.9/km 2  (546.2/sq mi). \bigstrut[t] \\
{\bf Question:} Curitiba is above sea level. True or False?\\
{\bf Answer:} The elevation of Curitiba are 934.6 m (3,066.3 ft). Elevation is a hypernym of level. Thus, Curitiba is above sea level. {\bf So the answer is true.}\\
Charles (Prince of Wales) was Born on 14 November 1948 ( 1948-11-14 )   (age 70)  Buckingham Palace, London, England.  The Spouse of Charles (Prince of Wales) are Lady Diana Spencer ( m.   1981 ;  div.   1996 )  , and Camilla Parker Bowles  ( m.   2005 ).  The Issue of Charles (Prince of Wales) are Prince William, Duke of Cambridge , and Prince Harry, Duke of Sussex. \bigstrut[t] \\
{\bf Question:} Charles was born in 1948 and has been married twice. True or False?\\
{\bf Answer:} Charles (Prince of Wales) was Born on 14 November 1948. The Spouse of Charles (Prince of Wales) are Lady Diana Spencer ( m.   1981 ;  div.   1996 )  , and Camilla Parker Bowles  ( m.   2005 ). Married is related to spouse. Thus, Charles was born in 1948 and has been married twice. {\bf So the answer is true.}\\
The Born of Idris Elba are 6 September 1972  (age 46)   Hackney, London, England.  The Residence of Idris Elba are London.  The Other names of Idris Elba are DJ Big Driis, Big Driis the Londoner, Big Driis, and 7 Dub.  The Occupation of Idris Elba are Actor, producer, director, musician, and DJ. \bigstrut[t] \\
{\bf Question:} Idris Elba is an English entertainer. True or False?\\
{\bf Answer:} The residence of Idris Elba is London. English is related to London. The occupation of Idris Elba are actor, producer, director, musician, and DJ. Actor is a hyponym of entertainer. Musician is a hyponym of entertainer. DJ is an entertainer. Thus, Idris Elba is an English entertainer. {\bf So the answer is true.} \\
The Breed of Jean, the Vitagraph Dog are Scotch Collie.  The Sex of Jean, the Vitagraph Dog are Female.  The Born of Jean, the Vitagraph Dog are 1902 Eastport, Maine.  The Years active of Jean, the Vitagraph Dog are 1909 - 1916. \bigstrut[t] \\
{\bf Question:} Jean, the Vitagraph Dog was a Golden Retriever which perform in circus. True or False?\\
{\bf Answer:} The Breed of Jean, the Vitagraph Dog are Scotch Collie. Collie is a hyponym of dog. Retriever is a hyponym of dog. Thus, Jean, the Vitagraph Dog was not a Golden Retriever which perform in circus. {\bf So the answer is false.}\\
The Studio of Hydrograd are Sphere Studios, North Hollywood, Los Angeles.  The Genre of Hydrograd are Hard rock.  The Label of Hydrograd are Roadrunner.  The Producer of Hydrograd are Jay Ruston. \bigstrut[t]\\
{\bf Question:} Hydrograd is in the rap genre. True or False?\\
{\bf Answer:} The Genre of Hydrograd are Hard rock. Rap is distinct from rock. Thus, Hydrograd is not in the rap genre. {\bf So the answer is false.} \bigstrut[b]\\
\hline
\end{tabular}}
\caption{The CoT prompt for tabular reasoning.
}
\label{table:tabular-prompt}
\end{table*} 

\end{document}